\documentclass[10pt,twocolumn,letterpaper]{article}

\usepackage{cvpr}
\usepackage{graphicx}
\usepackage{amsmath,amssymb} 
\usepackage{color}
\usepackage{wrapfig,lipsum,booktabs}
\usepackage[ruled]{algorithm2e}
\usepackage{algpseudocode}
\usepackage{caption}
\usepackage{subfig}
\usepackage{multirow}

\usepackage[breaklinks=true,bookmarks=false]{hyperref}

\cvprfinalcopy 

\def\cvprPaperID{****} 

\ifcvprfinal\fi
\begin{document}

\title{Deep Structure Inference Network for Facial Action Unit Recognition}

\author{\parbox{16cm}{\centering
    {\large Ciprian A. Corneanu$^{1}$, Meysam Madadi$^{2,3}$, Sergio Escalera$^{1,2}$}\\
    {\normalsize
    $^1$ Dept. Mathematics and Informatics, Universitat de Barcelona, Catalonia, Spain\\
    $^2$ Computer Vision Center, Edifici O, Campus UAB, 08193 Bellaterra (Barcelona), Catalonia, Spain\\
    $^3$ Dept. of Computer Science, Univ. Aut\`onoma de Barcelona (UAB), 08193 Bellaterra, Catalonia, Spain\\
    }}
}

\maketitle
\thispagestyle{empty}

\begin{abstract}
   Facial expressions are combinations of basic components called Action Units (AU). Recognizing AUs is key for developing general facial expression analysis. In recent years, most efforts in automatic AU recognition have been dedicated to learning combinations of local features and to exploiting correlations between Action Units. In this paper, we propose a deep neural architecture that tackles both problems by combining learned local and global features in its initial stages and replicating a message passing algorithm between classes similar to a graphical model inference approach in later stages. We show that by training the model end-to-end with increased supervision we improve state-of-the-art by 5.3\% and 8.2\% performance on BP4D and DISFA datasets, respectively.
\end{abstract}


\section{Introduction}
\label{sec:intro}

Facial expressions represent one of the most important cues for recognizing non-verbal behaviour, being an important part of human social signalling \cite{frith2009role}. The ability to automatically mine human intentions, attitudes or experiences has many potential applications like building socially aware systems \cite{vinciarelli09,devault2014}, improving e-learning experiences \cite{kapoor2007automatic}, adapting game status according to player's emotional responses \cite{bakkes2012personalised}, detecting pain for monitoring patients \cite{lucey10,kaltwang2012continous} and detecting deception during police interrogations or job interviews \cite{kulkarni2017automatic} just to name a few.  

\begin{figure*}%
    \centering
    \subfloat[]{{\includegraphics[height=0.37\linewidth]{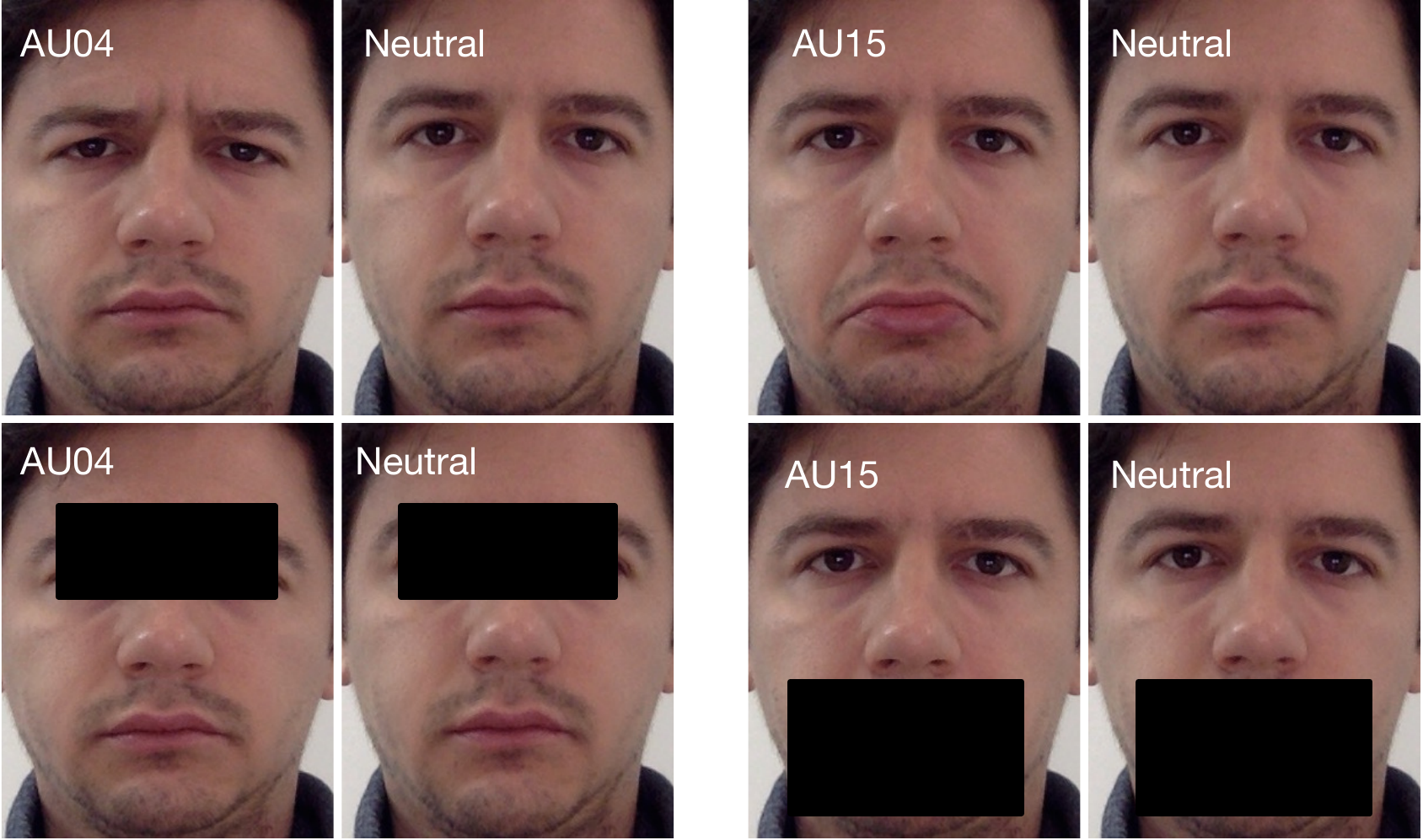} }\label{fig:motivation-local}}%
    \qquad
    \subfloat[]{{\includegraphics[height=0.37\linewidth]{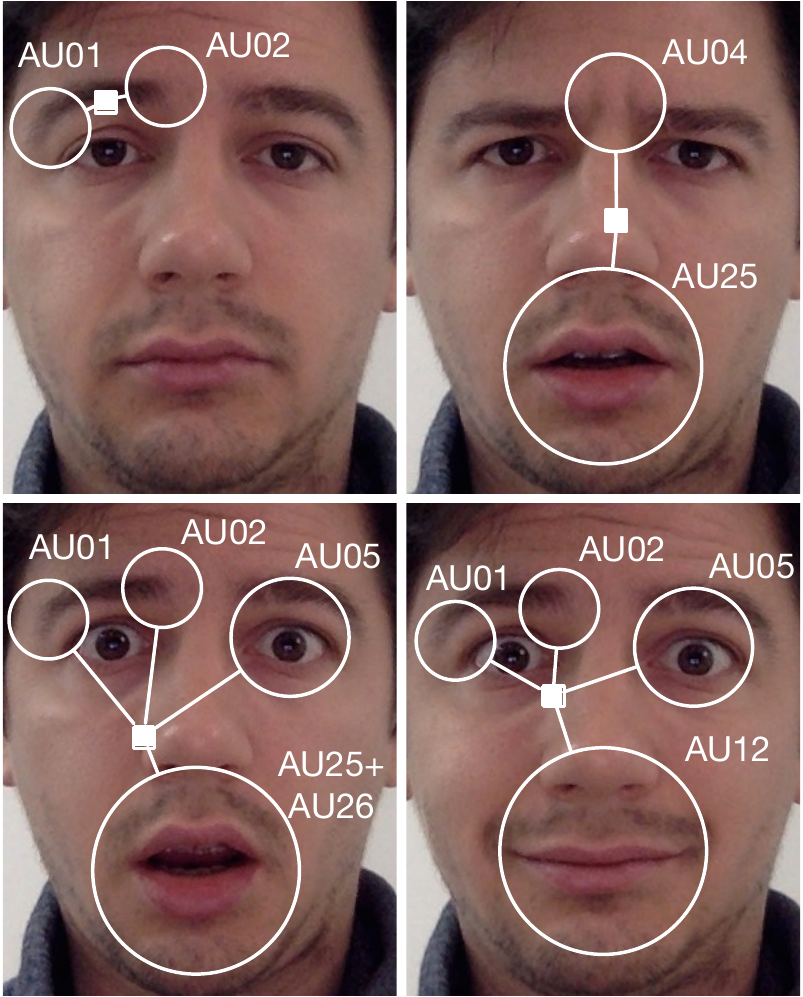} }\label{fig:motivation-multilabel}}
    \caption{AU patch and multi-label learning. (a) By masking just a small region an expressive face becomes indistinguishable from neutral. (b) Several AUs can be active at the same time and AU pairs can be strongly correlated.}
    \label{fig:motivation}
\end{figure*}

Result of the seminal research of Paul Ekman, the Facial Action Unit System (FACS) \cite{ekman2002facial}, is a descriptive coding scheme of facial expressions that focuses on what the face can do without assuming any cognitive or emotional value. Its basic components are called Action Units (AU) and they combine to form a complete representation of facial expressions. 
AUs are patterns of muscular activation and the way they modify facial morphology is relatively localized on the face (Fig. \ref{fig:motivation-local}). While initial methods for AU recognition (like JPML \cite{zhao2015joint} and APL \cite{zhong2015learning}) were using shallow predefined representations, recent methods (like DRML \cite{zhao2016deep}, ROI \cite{li2017action} and GL \cite{benitez2017recognition}) applied deep learning to learn richer local features that capture facial morphology. Therefore one could predict specific AUs from informative face regions adaptively selected depending on the facial geometry. For instance, contrary to non-adaptive methods like DRML \cite{zhao2016deep} and APL\cite{zhong2015learning}, ROI \cite{li2017action} and JPML \cite{zhao2015joint} extract features around facial landmarks which are more robust with respect to non-rigid shape changes. 
Patch learning is challenging as the human face is highly articulated and different face patches can contribute to either specific or groups of AUs. Learning the best patch combination together with learning specific features from each patch could be beneficial for AU recognition.     

Another key characteristic of AU recognition is that it is multi-label. Several AUs can be active at the same time and certain AU combinations are more probable than others (Fig. \ref{fig:motivation-multilabel}). Taking into account such probabilistic dependencies, AU prediction performance could be improved. As in deep approaches, such correlations can be addressed implicitly in the fully connected layers (e.g. DRML \cite{zhao2016deep}, GL \cite{benitez2017recognition} and ROI \cite{li2017action}). However, structure is not learned in any explicit way and inference and sparsity are implicit by design. JPML \cite{zhao2015joint} treats the problem by including pre-learned priors about AU correlations into their learning. Learning structured outputs has been studied by \cite{zhao2015joint,walecki2017deep,eleftheriadis2015multi} using Graphical Models, mathematical models able to capture complex probabilistic high order inter-dependencies. However, these models are not end-to-end trainable. 

In this work, we claim patch and the multi-label learning are key problems in dealing with AU recognition. We propose a deep neural network that tackles those problems in an integrated way through an incremental and end-to-end trainable approach. First, the model learns local and holistic representations exhaustively from facial patches. Then it captures structure between patches by predicting specific AUs. Finally, AU correlations are captured by a structure inference network that replicates message passing inference algorithms in a connectionist fashion. Tab. \ref{tab:comparison} compares some of the most important features of the proposed method to the state-of-the-art (specifically JPML \cite{zhao2015joint}, APL \cite{zhong2015learning}, DRML\cite{zhao2016deep}, GL\cite{benitez2017recognition} and ROI\cite{li2017action}). We show that by separately treating problems in different parts of the network and being able to optimize them jointly, we improve state-of-the-art by 5.3\% and 8.2\% performance on BP4D and DISFA datasets, respectively. Summarizing, our 2 main contributions are: 1) we propose a model that learns representation, patch and output structure end-to-end, and 2) we introduce a structure inference topology that replicates inference algorithm in probabilistic graphical models by using a recurrent neural network.

\begin{table}[t]
    \centering
        \begin{tabular}{|c|c|c|c|c|c|}
        \hline
        method & LRL & AP & PL & SL & EE \\ \hline
        APL \cite{zhong2015learning} & $\times$ &  $\times$  & $\checkmark$  & $\times$  & $\times$  \\ \hline
        GL \cite{benitez2017recognition}   & $\times$ &  $\times$  & $\checkmark$  & $\times$  & $\checkmark$ \\ \hline
         
        JPML \cite{zhao2015joint}  & $\times$ &  $\checkmark$  & $\checkmark$ & $\times$  & $\times$ \\ \hline
        ROI \cite{li2017action} & $\checkmark$ &  $\checkmark$  & $\checkmark$  & $\times$  & $\checkmark$ \\ \hline
         
        DRML \cite{zhao2016deep}  & $\checkmark$ &  $\checkmark$  & $\times$  & $\times$  & $\checkmark$ \\ \hline
        DSIN (ours) & $\checkmark$ &  $\checkmark$  & \textbf{$\checkmark$} & \textbf{$\checkmark$}  & \textbf{$\checkmark$}\\\hline
        \end{tabular}
        \caption{\small Features of our model and related work. LRL: local representation learning, AP: adaptive patch, PL: patch learning, SL: structured learning, EE: end-to-end.}
    \label{tab:comparison}
\end{table}

The paper is organised as follows. Sec. \ref{sec:related_work} presents related work. Sec. \ref{sec:method} details the proposed model and Sec.  \ref{sec:experimental_analysis} the results. Sec. \ref{sec:conclusion} concludes the paper.

\section{Related Work}
\label{sec:related_work}


In this section we group related work in three main categories: patch learning, multi-label and structure learning, and other works that focus on alternative problems like fusing temporal features or representation learning.

\textbf{Patch Learning.} 
Inspired by locally connected convolutional layers~\cite{taigman2014deepface}, Zhao et al. \cite{zhao2016deep} proposed an intermediate regional connected convolutional layer that learns specific convolutional filters from sub-areas of the input, showing improvements over standard convolutional layers. In \cite{li2017action}, different CNNs are trained on different parts of the face merging features in an early fusion fashion with fully connected layers. Zhao et al. \cite{zhao2015joint} performed patch selection and structure learning with shallow representations where patches for each AU were selected by group sparsity learning. Jaiswal et al. \cite{jaiswal2016deep} used domain knowledge and facial geometry to pre-select a relevant image region for a particular AU, passing it to a convolutional and bi-directional Long Short-Term Memory (LSTM) neural network. Zhong et al. \cite{zhong2015learning} proposed a multi-task sparse learning framework for learning common and specific discriminative patches for different expressions. Patch location was predefined and did not take into account facial geometry.

\textbf{Multi-label and Structure Learning.} Zhang et al. \cite{zhang2016task} proposed a multi-task approach to learn a common kernel representation that describes AU correlations. Elefteriadis et al. \cite{eleftheriadis2015multi} adopted a latent variable Conditional Random Field (CRF) to jointly detect multiple AUs from predesigned features. While existing methods capture local pairwise AU dependencies, Wang et al. \cite{wang2013capturing} proposed a restricted Boltzmann machine that captures higher-order AU interactions. Together with patch-learning, Zhao et al. \cite{zhao2015joint} used positive and negative competitions among AUs to model a discriminative multi-label classifier. Walecki et al. \cite{walecki2017deep} placed a CRF on top of deep representations learned by a CNN. Both components are trained iteratively to estimate AU intensity.  Wu et al. \cite{wu2016constrained} used a Restricted Boltzman Machine that captures joint probabilities between facial landmark locations and AUs. More recently, Benitez et al. \cite{benitez2017recognition} proposed a loss combining the recognition of isolated and groups of AUs.

\textbf{Others.} Alternative methods have looked into fusing multi-scale temporal  decisions or learning separable representations. Ding et al. \cite{ding2013facial} proposed the fusion of complementary classifiers on frame-level detection, segment-level detection (detecting AU segments from contiguous frames) and transition detection (recognizing transitions between AU and non-AU frames). Zeng et al. \cite{zeng2015confidence} proposed an ensemble of classifiers that makes decisions in an easy-to-hard fashion using predefined shallow features extracted around facial landmarks. 
Song et al. \cite{song2015exploiting} proposed a feature disentangling machine that is capable of selecting features into non-overlapped groups. They focus on common features that are shared across different expressions and expression-specific features that are discriminative only for a target expression. 

Contrary to the above we propose a model capable of learning deep local representations, patches and structure jointly. We do this by implementing a message passing inference algorithm using a recurrent neural topology.

\section{Method}
\label{sec:method}
Let $\mathcal{D}=\{\mathbf{X},\mathbf{Y}\}$ be a set of pairs of input images $\mathbf{X}=\{\mathbf{x}_1,...,\mathbf{x}_M\}$ and output AU labels $\mathbf{Y}=\{\mathbf{y}_1,...,\mathbf{y}_M\}$ with $M$ number of instances. Each input image $\mathbf{x}_i$ is composed of $P$ image patches $\{I_1,...,I_P\}$ and output label $\mathbf{y}_i$ is a set of $N$ AUs $\{y_1,...,y_N\}$, each $y_j$ taking a binary value $\{0,1\}$. This means several AU classes can be active for an observation as a multi-label problem. Predicting such output is challenging as a softmax function can not be applied on the set of outputs contrary to the standard mono-label/multi-class problems. In addition, using independent AU activation functions in losses like cross-entropy, ignores AU correlations. Including the ability to learn structure in the model design is thus of key relevance.

Two main ways of solving multi-label learning in AU recognition are either capturing correlations through fully-connected layers \cite{zhao2016deep,benitez2017recognition,li2017action} or inferring structure through probabilistic graphical models (PGM) \cite{zhao2015joint,walecki2017deep,eleftheriadis2015multi}. While the former can capture correlations between classes, this is not done explicitly. On the other hand, PGMs offer an explicit solution and their optimization is well studied. Unfortunately, placing classical PGMs on top of neural network predictions considerably lowers the capacity of the model to learn high order relationships since it is not end-to-end trainable. One solution is to replicate graphical model inference in a conectionist fashion which would make possible joint optimization. Jointly training CNNs and CRFs has been previously studied in different problems \cite{zheng2015conditional,chu2016crf,deng2016structure}. Following this trend, in this work we formulate AU recognition by a graphical model and implement it by neural networks, more specifically CNNs and recurrent neural network (RNN). This way, AU predictions from local regions along AU correlations are learned end-to-end.

\begin{figure*}[t]
\begin{center}
   \includegraphics[width=0.9\textwidth]{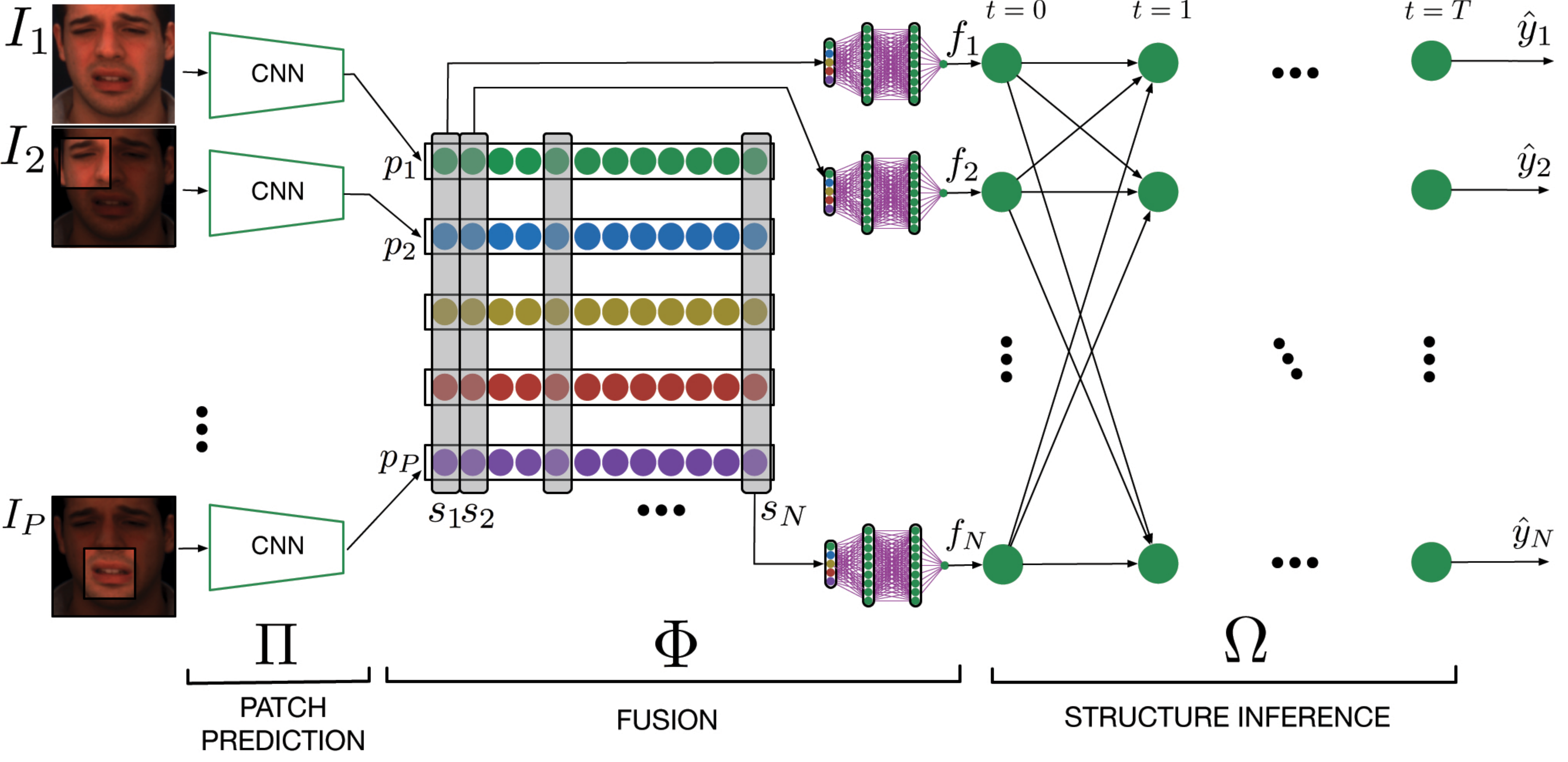}
\end{center}
   \caption{\small Deep Structure Inference Network (DSIN). DSIN learns independent AU predictions from global and local deeply learned features and replicates a message passing mechanism between AUs. It refines each AU prediction by taking into account correlation to the other AUs. Each input image is cropped into a set of patches $\{I_i\}_{i=1}^{P}$ which is used for training an independent CNN for producing a probability vector $p_i$ for $N$ AUs ($\varphi_p$ in Eq. \ref{eq:crf_energy}). From $s_j$ (the patch predictions for a specific AU) we learn a combination for producing a single AU prediction $f_j$ (simplified $\psi_{py}$ in Eq. \ref{eq:crf_energy}). Final predictions $y_j$ are computed by inferring structure among AUs through iterative message passing similar to inference in a probabilistic graph model ($\psi_y$ in Eq. \ref{eq:crf_energy}).}
\label{fig:method}
\end{figure*}

Let $\mathcal{G}=(\mathcal{V},\mathcal{E})$ denote a graph with vertices $\mathcal{V}=\mathbf{y}$ specifying AUs and edges $\mathcal{E}\subseteq \mathcal{V}\times \mathcal{V}$ indicating the relationships between AUs. Given the Gibbs distribution we compute conditional probability $\mathnormal{P}(\mathbf{y}|\mathbf{x},\Theta)$ as:
\begin{equation}
    \mathnormal{P}(\mathbf{y}|\mathbf{x},\Theta)=\frac{1}{Z(\mathbf{y},\mathbf{x},\Theta)}e^{-E(\mathbf{y}|\mathbf{x},\Theta)},
\end{equation}
where $\Theta$ are model parameters, $Z$ is a normalization function and $E$ is an energy function. The model can be updated by introducing some latent variables $\mathbf{p}$ as:
\begin{equation}
    \mathnormal{P}(\mathbf{y}|\mathbf{x},\Theta)=\sum_\mathbf{p} \mathnormal{P}(\mathbf{y},\mathbf{p}|\mathbf{x},\Theta),
\end{equation}
where $\mathbf{p}$ is given as the output of CNN. Therefore, the vertices and edges in the graph $\mathcal{G}$ can be updated as $\mathcal{V}=\mathbf{y} \cup \mathbf{p}$ and $\mathcal{E}=\mathcal{E}_y \cup \mathcal{E}_{py} \cup \mathcal{E}_p$. Although edges $\mathcal{E}_y$ can be defined by a prior knowledge taken from a given dataset, we use a fully connected graph independent to the dataset and assign a mutual gating strategy to control information passing through edges (more details in Sec. \ref{sec:sec:structure_inference}). We define $\mathcal{E}_{py}$ as edges between $\mathbf{p}$ and $\mathbf{y}$, and use a selective strategy to define edges in this set. Finally, edges $\mathcal{E}_p$ is an empty set, since in our model an independent CNN is trained on each image patch $I_j$ and we do not assign any edge among $\mathbf{p}$. Given this assumption, probability distribution $\mathnormal{P}(\mathbf{y},\mathbf{p}|\mathbf{x},\Theta)$ is given by:
\begin{equation}
    \mathnormal{P}(\mathbf{y},\mathbf{p}|\mathbf{x},\Theta)=\mathnormal{P}(\mathbf{y}|\mathbf{p},\mathbf{x},\Theta) \prod_{k} \mathnormal{P}(p_k|\mathbf{x},\Theta).
\end{equation}

As in CRF, energy function $E(.)$ is computed by unary and pairwise terms as:
\begin{equation}\small
\begin{split}
    E(\mathbf{y},\mathbf{p},\mathbf{x},\Theta)= & \sum_k \varphi_p(p_k,\mathbf{x},\pi) + \sum_{(i,k)\in \mathcal{E}_{py}} \psi_{py}(y_i,p_k,\phi) \\
    & + \sum_{(i,j)\in \mathcal{E}_y} \psi_y(y_i,y_j,\omega),
    \end{split}
    \label{eq:crf_energy}
\end{equation}
where $\varphi(.)$ is a unary term, $\psi_*(.)$ are pairwise terms and $\Theta=\pi \cup \phi \cup \omega$. Fig. \ref{fig:method} presents our Deep Structure Inference Network (DSIN). It consists of three components each designed to solve a term in Eq. \ref{eq:crf_energy}. We refer to the initial part as \emph{Patch Prediction} (PP), whose purpose is to exhaustively learn deep local representations from facial patches and produce local predictions. Then, the \emph{Fusion} (F) module performs patch learning per AU. The final stage, \emph{Structure Inference} (SI), refines AU prediction by capturing relationships between AUs. The DSIN is end-to-end trainable and CNN features can be trained based on gradients back-propagated from structure inference in a multi-task learning fashion.

\begin{figure*}[ht!]
    \centering
    \includegraphics[width=0.9\textwidth]{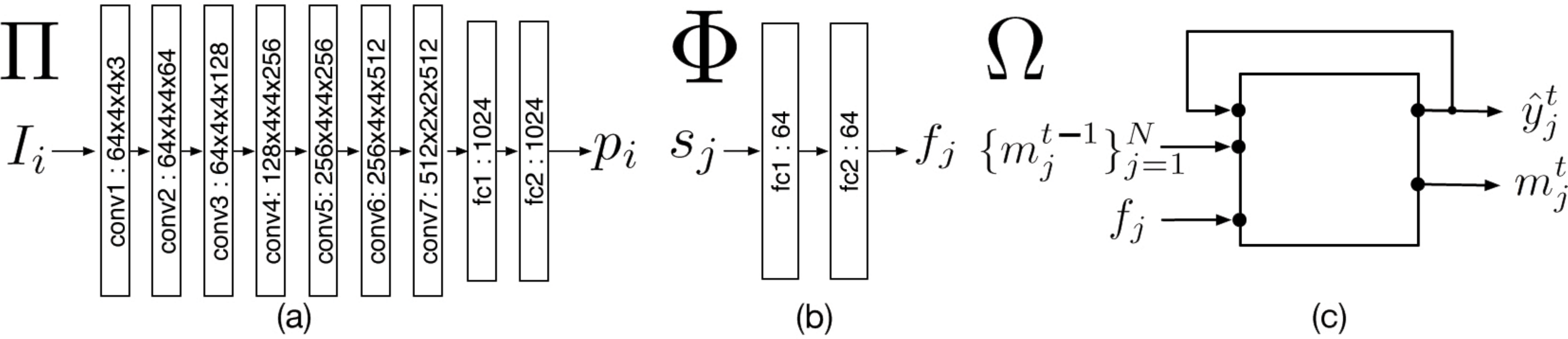}
    \caption{\small (a) Topology of patch prediction CNNs. Each convolutional block consists of a convolutional layer with stride 2 and batch normalization. The convolutional layer is shown by the number of filters followed by the size of the kernel. The last layers are fully-connected (FC) layers marked with the number of neurons. All neurons use ReLU activation functions. (b) Each fusion unit is a stack of 2 FC layers. (c) A structure inference unit. For better visualization, we just show the interface of the unit without the inner topology. See details in Sec. \ref{sec:sec:structure_inference}.}
    \label{fig:method:details}
\end{figure*}

\subsection{Patch Prediction} 
Given image patches $\mathbf{x}$, unary terms $\varphi_p(\mathbf{p},\mathbf{x},\pi)$ provide AUs confidences for each patch which are defined as the log probability:
\begin{equation}
    \varphi_p(\mathbf{p},\mathbf{x},\pi)=\log \mathnormal{P}(\mathbf{p}|\mathbf{x},\pi).
\end{equation}
Probability $\mathnormal{P}(\mathbf{p}|\mathbf{x},\pi)$ is modeled by independent functions, the patch prediction functions $\{\Pi_i(I_i;\pi_i)\}_{i=1}^{P}$, where $I_i$ is input image patch and $\pi_i$ are function parameters. Each $\Pi_i$ is a CNN providing $N$ AUs probabilities through sigmoid function at last layer. $P$ independent predictions are provided at this stage, each being a vector of AU predictions. Although image patches may have overlap, we choose independence assumption to let each network to be expert at predicting AUs on local regions. By learning independent global representations from the whole face and local representations from sub-regions of the face, we can better capture facial morphology and better address AU locality. 

In Fig. \ref{fig:method:details}(a) we detail the topology of the CNNs used for learning the patch prediction functions. Many complex convolutional topologies have been proposed in recent years and searching for the best topology is out of the scope of this work. The chosen topology, which is a shallow network, follows the intuition behind well known models like VGG~\cite{simonyan2014very}. 

\subsection{Fusion} 
Computational complexity to marginalize pairwise relationships in $\mathcal{E}_{py}$ is high. In our formulation, we simplify edges such that $\mathcal{E}_{py}$ becomes directed from nodes in $\mathbf{p}$ to nodes in $\mathbf{y}$. It means we omit mutual relationships among $\mathbf{p}$ and $\mathbf{y}$. Therefore, nodes in $\mathbf{y}$ are conditioned on the nodes in $\mathbf{p}$. However, we want each AU node in $\mathbf{y}$ to be conditioned on the same AU nodes in $\mathbf{p}$ from different patches. It means different patches can provide complementary information to predict target AU independent to other AUs. Finally, $\psi_{py}(\mathbf{y},\mathbf{p},\phi)$ is defined as the log probability of $\mathnormal{P}(\mathbf{y}|\mathbf{p},\phi)$ which is modeled by a set of independent functions, so called fusion functions $\{\Phi_j(s_j;\phi_j)\}_{j=1}^N$, where $s_j\subset \mathbf{p}$ corresponds to the set of $j$-th AU predictions from all patches and $\phi_j$ is function parameters. We simply model each function $\Phi_j$ with 2 fully connected layers with 64 hidden units, each followed by a sigmoid layer, as shown in Fig. \ref{fig:method:details}(b). We found 64 hidden units works well in practice while higher dimensionality does not bring any additional performance and quickly starts over-fitting. The output of each $\Phi_j$ is the predicted probability $f_j$ for $j$-th AU.

\subsection{Structure Inference}
\label{sec:sec:structure_inference}
Up to now, we computed individual AU probabilities in a feed-forward neural network without taking AU relationships explicitly into account. The goal is to model pairwise terms $\psi_y$ such that the whole process is end-to-end trainable in a compact way. Belief propagation and message passing between nodes is one of the well known algorithms for PGM inference. Inspired by \cite{deng2016structure}, which proposes a connectionist implementation for action recognition, we build a \emph{Structure Inference} (SI) module in the final part of DSIN. 

The SI updates each AU prediction in an iterative manner by taking into account information from other AUs. The intuition behind this is that by passing information between predictions in an explicit way, we can capture AU correlations and improve predictions. The structure inference module is a collection of interconnected recurrent structure interference units (SIU) (see Fig. \ref{fig:method:details}(c)). For each AU there is a dedicated SIU. We denote the computations done by SIU by a function $\Omega$. Let $\{\Omega_j\}_{j=1}^N$ be the set of SIU functions $\Omega_j:\mathbb{R}^{N+2} \rightarrow \mathbb{R}^2$ where:
\begin{equation}
    \hat{y}_j^{t} = \Omega_j(f_j, m_1^{t-1}, m_2^{t-1}, ..., m_N^{t-1}, \hat{y}_j^{t-1};\omega_j).
\end{equation}
At each iteration $t$, $\Omega_j$ takes as input the initial prediction $f_{j}$ for its class, a set of incoming messages $\{m_j^{t-1}\}_{j=1}^{N}$ from the SIUs corresponding to the other classes and its own previous prediction $\hat{y}_j^{t-1}$. Each function $\Omega_j$ has two inline units: producing $j$-th AU prediction $\hat{y}_j^{t}$ and message $m^{t}_{j}$ for next time step. In this way, predictions are improved iteratively by receiving information from other nodes. Computationally, we replicate this iterative message passing mechanism in the collection of SIUs with a recurrent neural network that shares function parameters $\Omega_j$ across all time steps. We show a SIU unit in Fig. \ref{fig:method:details}(c).

A message unit basically corresponds to the distribution of the AU node. A message unit from a SIU is a parametrized function of the previous messages, the initial fused prediction and the previous prediction of the same SIU:
\begin{equation}
    m_{j}^{t} = \sigma\left(\omega_{j}^{m}\left[\mu(m_{1}^{t-1}, ..., m_{N}^{t-1}), f_j, \hat{y}_j^{t-1}\right] + \beta_{j}^{m} \right),
\end{equation}
where $\sigma(.)$ is the sigmoid function, $\mu(.)$ is the mean function, $\omega_{j}^{m}\in\mathbb{R}^3$ and $\beta_{j}^{m}\in\mathbb{R}$ are message function parameters. Messages between two nodes at each time step have a mutual relationship which can be controlled by a gating strategy. Therefore, a set of correction factors are computed as:
\begin{equation}
    \chi_j^{t} = \sigma\left(
        \omega_{j}^{g} \left[\mu(m_{1}^{t}, ..., m_{N}^{t}), f_j,  \hat{y}_j^{t-1}\right] + \beta_j^g  \right),
    \label{eq:correction_factor}
\end{equation}
where $\omega_{j}^{g}\in\mathbb{R}^3$ and $\beta_{j}^{g}\in\mathbb{R}$ are gating function parameters. Then, a message $m_{i\rightarrow j}^{t}$ that is passed from AU node $i$ to $j$ will be updated by the mutual factors of the gate between nodes $i$ and $j$ as:
\begin{equation}
    \overline{m}^{t}_{j} = \mu(\chi_{i}^{t}, \chi_{j}^{t}) m_{i\rightarrow j}^{t}.
    \label{eq:corrected_message}
\end{equation}
Finally, updated messages coming to the $j$-th node along with initial estimation $f_j$ are used to produce output prediction $\hat{y}_j^{t}$ as:
\begin{equation}
    \hat{y}_j^{t} = \sigma\left(
        \omega_{j}^y \left[\mu(\overline{m}_{1}^{t}, ..., \overline{m}_{N}^{t}),  f_j\right] + \beta_j^{y}    \right),
    \label{eq:final_prediction}
\end{equation}
where $\omega_{j}^{y}\in\mathbb{R}^2$ and $\beta_{j}^{y}\in\mathbb{R}$ are prediction function parameters. By doing this, we are able to combine representation learning in function $\Pi$, patch learning in function $\Phi$ and structure inference in the $\Omega$ in a single end-to-end trainable model. We introduce our training strategy in Sec. \ref{sec:training}.

\begin{figure}[ht!]
\begin{center}
   \includegraphics[width=\linewidth]{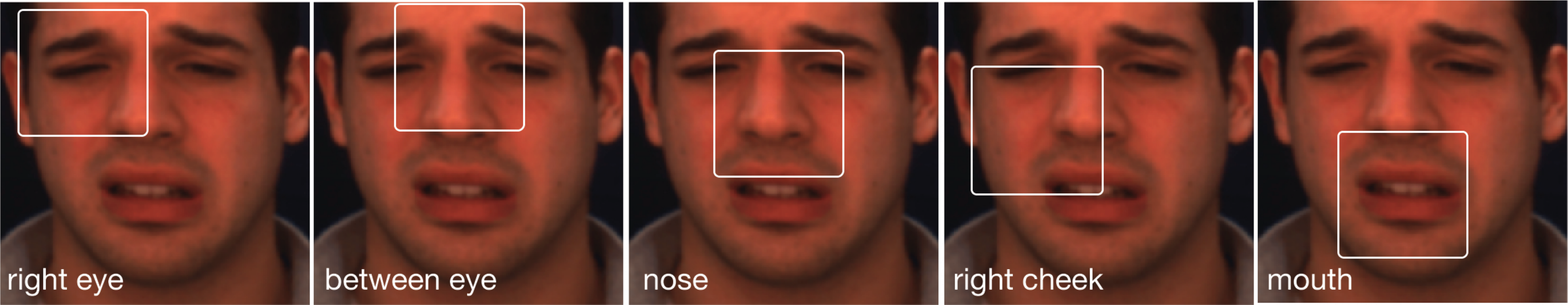}
\end{center}
   \caption{Each input image is aligned and cropped into 5 patches.}
\label{fig:patches}
\end{figure}

\section{Experimental Analysis}
\label{sec:experimental_analysis}
This section describes the experimental setting and presents detailed results.

\subsection{Experimental Setting}
Here, we detail datasets, preprocessing, and evaluation metrics and  methods.  

\begin{table}[t]
    \centering
    \begin{tabular}{|c|c|c|}
        \hline 
        & DISFA\cite{mavadati2013disfa}  & BP4D\cite{zhang2014bp4d}  \\\hline
        \#seqs & 27 & 328 \\\hline
        \#frames & 130,814 & 144,682 \\\hline
        \#active frames & 56,356 & 117,075 \\\hline
        \#AU & 10 & 12 \\\hline
        label cardinality$^\dag$ & 3.04 & 4.05 \\\hline
        label density$^{\ddag}$ & 4.05 & 0.22 \\\hline
    \end{tabular}
    \caption{\small Datasets used. $^{\dag}$ average number of labels per observation. $^{\ddag}$ number of labels per observation divided by the total number of labels, averaged over the samples.}
    \label{tab:datasets}
\end{table}

\begin{figure}[t]
\centering
\includegraphics[width=0.8\linewidth]{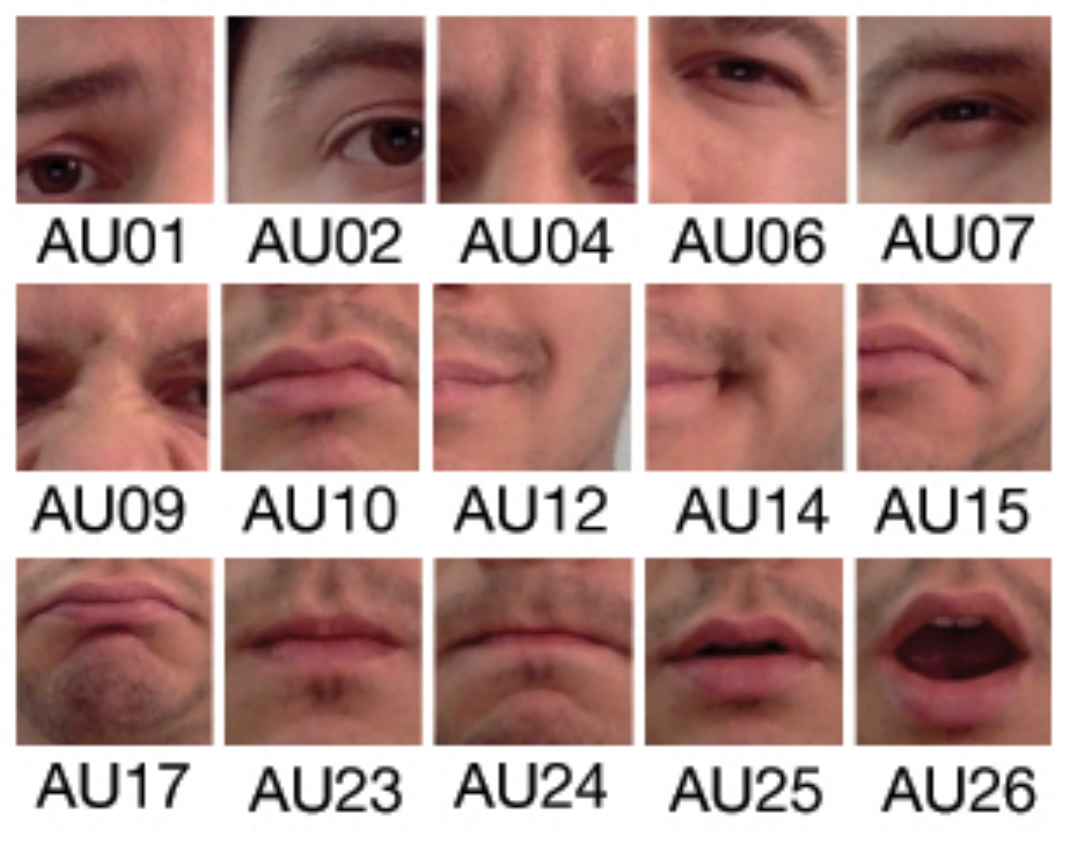}
\captionof{figure}{\small Facial Action Units targeted in this work.}
\label{fig:aus}
\end{figure}

\subsubsection{Data.}
\label{sec:experimental_analysis:data}
We used BP4D \cite{zhang2014bp4d} and DISFA \cite{mavadati2013disfa} datasets. BP4D contains 2D and 3D videos of 41 young adults during various emotion inductions while interacting with an experimenter. It has 328 videos (8 videos for 41 participants) with 12 coded AUs, resulting in about 140k valid face images \cite{zhang2014bp4d}.  DISFA contains 27 adults (12 women and 15 men) with ages between 18 to 50 years and relative ethnic diversity. Participants viewed a 4-minute video clip (242 seconds in length) intended to elicit spontaneous AUs. The data corpus consists of approximately 130k frames in total. AU intensity was coded for each video frame on a 0 (not present) to 5 (maximum intensity) ordinal scale. For our purpose we consider all labels with intensity greater than 3 as active and the rest as non-active. Table \ref{tab:datasets} shows datasets characteristics. Overall, BP4D has considerable higher label density (more active AU per frame) and greater number of sequences of shorter length. Both datasets are considered in most recent works for AU recognition.

\begin{algorithm*}[th!]
    \textbf{Training data}: $\{ \{I\}_{i=1}^{P},y\}$\\
    \textbf{Model parameters}: patch prediction: $\{\pi_i\}_{i=1}^{P}$, fusion $\{\phi_i\}_{i=1}^{N}$, structure inference $\{\omega_i\}_{i=1}^{N}$\\
    
    \textbf{Step 0:} random initialization around 0: $\pi,\phi,\omega \gets \mathcal{N}(0,\sigma^2)$ \\
    
    \textbf{Step 1:} train patch prediction: 
    $ \pi_i \gets \min_{\pi}  (L_{\Pi}(\Pi_i(I_i;\pi_i)),y), \forall i \in \{1,...,P\}$
        
    \textbf{Step 2:} freeze patch prediction; train fusion: $\phi \gets \min_{\phi} L_{\Phi}(\Phi(\Pi;\phi),y)$
    
    \textbf{Step 3:} train patch prediction and fusion jointly:\\
    
    \hspace{.7 cm} $\pi,\phi \gets \min_{\pi, \phi} ( 
    L_{\Pi}(\Pi(I;\pi)),y) + L_{\Phi}(\Phi(\Pi;\phi),y)
    )$ 

    \textbf{Step 4:} freeze patch prediction and fusion; train structure inference:\\
    
    \hspace{.7cm} $\omega \gets \min_{\omega} L_{\Omega}(\Omega(\Phi; \omega),y)$   
    
    \textbf{Step 5.} train all: \\
    $\pi, \phi, \omega \gets \min_{\pi, \phi, \omega} ( 
    w_1L_{\Pi}(\Pi(I;\pi)),y)+
    w_2L_{\Phi}(\Phi(\Pi;\phi),y)+
    w_3L_{\Omega}(\Omega(\Phi; \omega),y) )$ 
    
    \textbf{Output}: optimized parameter: $\pi^{opt}$, $\phi^{opt}$, $\omega^{opt}$
 \caption{Training procedure of DSIN.}
 \label{algo:training}
\end{algorithm*}


\subsubsection{Preprocessing.} For each image, facial geometry is estimated by an ensemble of regression trees \cite{kazemi2014one}. From geometry of neutral faces of all subjects we compute 3 reference anchors by the mean of all landmarks corresponding to the two eyes and the mouth. Faces are resized to $224 \times 224 \times 3$ and a rigid transformation is applied for registering to anchors, reducing variance to scale and rotation. We crop 5 patches of size $56 \times 56 \times 3$ around interest points defined by the detected landmarks (see Fig. \ref{fig:patches}). The 5 patches cover relevant parts of the face like the eyes, mouth or nose. For reducing redundancy we ignore patches on the corresponding symmetrical part of the face like the left eye and cheek.

\subsubsection{Training.}\label{sec:training} We incrementally train each part of DSIN before end-to-end model training. During training we use supervision on the patch prediction $p$, the fusion $f$ and the structure inference outputs $\hat{y}$. On $p$ we use a weighted $L_2$ loss denoted by $L_{\Pi}(p,y)$. The weights are inversely proportional to the ratio of positives in the total number of observations for each AU class in training. The weighting gives more importance to the minority classes in each training batch which ensures a more equal gradient update across classes and overall better performance. On the fusion and structure inference outputs we apply a binary cross-entropy loss (denoted by $L_\Phi(f,y)$ and $L_\Omega(\hat{y},y)$). For the structure inference we include a regularization on the correction factors (denoted by $\chi$ in Eq. \ref{eq:correction_factor} and Eq. \ref{eq:corrected_message}) to force sparsity in the message passing. Details of the training procedure are shown in Alg. \ref{algo:training}. For training we use an Adam optimizer with learning rate of 0.001 and mini-batch size 64 with early stopping. Experimentally, we found the individual loss contributions $w_1=0.25$, $w_2=0.25$ and $w_3=0.5$ in the final compound loss to work well in training. Sec. \ref{sec:experimental_analysis:ablation} presents an analysis of the effect of the correction factors regularization parameter $r$.

For both BP4D and the DISFA we perform a subject exclusive 3-fold cross-validation. Similarly to \cite{li2017action}, on DISFA we take the best CNNs trained for patch prediction on the BP4D and retrained fully connected layers for the new set of outputs. We fix the convolutional filters throughout the rest of the training. 

\subsubsection{Methods and metrics.} We compare against CPM \cite{zeng2015confidence}, APL \cite{zhong2015learning}, JPML \cite{zhao2015joint}, DRML \cite{zhao2016deep}, and ROI \cite{li2017action} state-of-the-art alternatives. We evaluate $F1$-frame score as $F1 = 2\frac{PR}{P+R}$, where $P=\frac{tp}{tp+fp}$, $R=\frac{tp}{tp+fn}$, $tp$ being true positives, $fn$ false negatives and $fp$ false positives. All metrics are computed per AU and then averaged. Targeted AUs shown in Fig.~\ref{fig:aus}.


\subsection{Results}
Here, in Sec. \ref{sec:experimental_analysis:ablation} we explore the effect of the design decisions included in the DSIN followed by comparison against state-of-the-art alternatives in Sec. \ref{sec:experimental_analysis:sota}. We conclude with a set of qualitative examples in Sec. \ref{sec:experimental_analysis:qualitative}.

\subsubsection{Ablation Study.}
\label{sec:experimental_analysis:ablation}
We analyze DSIN design decisions in the following.

\textbf{Class balancing.} In both BP4D and DISFA, classes are strongly imbalanced. This can be harmful during training. To alleviate this, we use a weighted loss on patch prediction CNNs. Tab. \ref{tab:ablation:bp4d} shows results with and without class balancing. This overall improves performance, especially on poorly represented classes. On BP4D the classes with ratios of positives in the total of samples lower than $30\%$ are AU01, AU02, AU04, AU17, AU24. These are the classes that are improved the most. AUs like AU07 or AU12 have positives to total rations higher than $50\%$. Balancing can reduce performance on these classes.

\begin{table*}[ht!]
    \centering
    \resizebox{\textwidth}{!}{%
    \begin{tabular}{|c|c|c|c|c|c|c|c|c|c|c|c|c|c|c|}
        \hline
        & method & AU01 & AU02 & AU04 & AU06 & AU07 & AU10 & AU12 & AU14 & AU15 & AU17 & AU23 & AU24 & avg\\ \hline

        & VGG(face)$^{ft}$ & \textbf{35.2} & 31.2 & 25.4 & 73.1 & \textbf{72.1} & 80.1 & 59.2 & 35.1 & 32.1 & 52.3 & 26.1 & \textbf{36.2} & 46.5 \\ \cline{2-15} 
        & PP(face)$^{ncb}$ & 35.1 & \textbf{38.1} & \textbf{53.9} & \textbf{77.2} & 70.7 & \textbf{83.1} & \textbf{86.2} & \textbf{56.1} & \textbf{39.8} & \textbf{54.5} & \textbf{37.2} & 31.4 & \textbf{55.3} \\ \hline \hline
        
        \multirow{8}{*}{\rotatebox[origin=c]{90}{\textbf{PP}}} & PP(right eye)$^{ind}$ & \textbf{46.8} & \textbf{40.4} & 45.3 & 68.3 & 69.2 & - & - & - & - & - & - & - & - \\ \cline{2-15}
        & PP(mouth)$^{ind}$ & - & - & - & - & - & 78.6 & 82.0 & 54.2 & 38.6 & 54.7 & [39.3] & \textbf{43.3} & - \\ \cline{2-15}
        
        & PP(right eye) & 38.0 & [37.7] & 48.3 & 69.5 & 71.0 & 72.4 & 77.4 & 50.7 & 15.0 & 38.9 & 13.8 & 15.3 & 45.7 \\  \cline{2-15}
        & PP(between eye) & 41.7 & 34.8 & 45.9 & 64.9 & 65.5 & 72.1 & 73.9 & 54.9 & 19.7 & 33.9 & 13.9 & 7.0 & 44.0 \\\cline{2-15}
        & PP(mouth) & 12.4 & 7.3 & 22.4 & 75.5 & 70.5 & 78.9 & 81.3 & \textbf{66.2} & 35.8 & 59.6 & 37.6 & [42.8] & 49.3 \\\cline{2-15}
        & PP(right cheek) & 30.5 & 18.4 & 41.8 & 75.2 & 73.2 & 79.1 & 81.9 & [61.9] & 35.7 & 55.1 & 35.5 & 35.7 & 52.0 \\\cline{2-15}
        & PP(nose) & 41.6 & 28.4 & 46.4 & 71.1 & 70.5 & 78.8 & 78.0 & 57.1 & 21.3 & 43.7 & 34.0 & 20.3 & 49.3 \\\cline{2-15}
        & PP(face) & 43.8 & 37.5 & [54.9] & \textbf{77.4} & [71.2] & [79.2] & \textbf{84.0} & 56.6 & [39.7] & [59.7] & 39.2 & 39.5 & [56.9] \\ \cline{2-15}
        & PP+F & [44.8] & 35.8 & \textbf{57.1} & [76.7] & \textbf{74.3} & \textbf{79.6} & [83.7] & 56.6 & \textbf{41.1} & \textbf{61.8} & \textbf{42.2} & 40.1 & \textbf{57.8} \\\hline \hline
        \multirow{5}{*}{\rotatebox[origin=c]{90}{\textbf{DSIN}}} & DSIN$_{2}^{ncf}$ & 46.7 & 34.1 & \textbf{62.0} & 76.5 & \textbf{74.1} & [83.1] & 84.9 & 60.9 & 36.0 & 57.1 & \textbf{43.3} & 36.1 & 57.9 \\\cline{2-15}
        & DSIN$_{2}$ & 47.7 & 36.5 & 55.6 & 76.3 & [73.7] & 80.1 & 85.0 & 64.0  & [39.2] & 60.6 & [43.1] & 39.9 & 58.2 \\\cline{2-15}
        & DSIN$_{5}$ & [49.7] & 36.3 & 57.3 & \textbf{76.8} & 73.4 & 81.6 & 84.5 & [64.7] & 38.5 & [63.0] & 39.0 & 37.3 & 58.5 \\\cline{2-15}
        & DSIN$_{10}$ & \textbf{51.7} & [40.4] & 56.0 & 76.1 & 73.5 & 79.9 & [85.4] & 62.7 & 37.3 & 62.9 & 38.6 & [41.6] & [58.9] \\\cline{2-15}
        & DSIN$_{10}^{tt}$ & \textbf{51.7} & \textbf{41.6} & [58.1] & [76.6] & \textbf{74.1} & \textbf{85.5} & \textbf{87.4} & \textbf{72.6} & \textbf{40.4} & \textbf{66.5} & 38.6 & \textbf{46.9} & \textbf{61.7} \\\hline
    \end{tabular}%
}
    \caption{\small Recognition results on BP4D.  PP([patch]) stands for patch prediction on the indicated patch. F stands for the fusion and DSIN is the final model. We indicate the results when training on individual AUs with [method]$^{ind}$, fine tuning on the validation dataset of the decision threshold by DSIN$^{tt}$, number of iterations of the structure inference by DSIN$_{T}$ and training without correction factors as DSIN$^{ncf}$. VGG(face)$^{ft}$ is a pre-trained VGG-16 \cite{parkhi2015deep} fine-tuned on BP4D. PP(face)$^{ncb}$ is a patch prediction without class balancing. All results are obtained by 3-fold cross-validation on BP4D.}
    \label{tab:ablation:bp4d}
\end{table*}

\textbf{Choice of prediction topology.} In Tab. \ref{tab:ablation:bp4d} we compare the proposed CNN topology for patch prediction (PP(face)) against the well-known VGG-16. The VGG-16 model used was trained for face recognition \cite{parkhi2015deep} and fine-tuned on our data for AU recognition. Our proposed topology shows superior performance.

\textbf{Targeting subsets of AUs.} We explore the effect of the considered target set on the overall prediction performance. In Tab. \ref{tab:ablation:bp4d} we show prediction results from the right eye and from the mouth patches when training either on the full set of targets ($[method]$) or on individual targets ($[method]^{ind}$). When training on individual AUs the decision for the classifier is considerably simpler. On the other hand any correlation information between classes that could be captured by the fully connected layers is ignored. In certain cases the individual prediction is superior to the exhaustive prediction. In the case of the right eye patch this is particularly true for AU01. But this is rather the exception. On average and across patches training on groups of AUs or on all AUs is beneficial as correlation information between classes is employed by the network in the fully connected layers. Additionally, predicting AU individually with independent nets would quickly increase the number of parameters with considerable effects on the training speed and final model performance.

Tab. \ref{tab:ablation:bp4d} and \ref{tab:ablation:disfa} show AU recognition results on both datasets trained on patches. That proves the locality assumption. When training on the mouth the performance on the upper face AUs is greatly affected. Similarly, training on the eye affects the performance on the lower face AUs. This is expected as the patch prediction can only infer the other AUs from the ones visible in the patch.

\textbf{Learning Local Representations.} On average, face prediction compared to patch prediction performs better on the entire output set. However, when individual AUs are considered, this is no longer the case. For BP4D, the performance on AU15 and AU24 are considerably higher when predicting from the mouth patch than from the face (see Tab. \ref{tab:ablation:bp4d}). On DISFA the prediction from the whole face is the best on just 3 AUs (see Tab. \ref{tab:ablation:disfa}). The nose patch is better for predicting AU06 and AU09, the mouth patch is better for AU12, AU25 and AU26, and the between eye patch for AU01.

\begin{figure*}[ht!]
    \centering   
    \includegraphics[width=\linewidth]{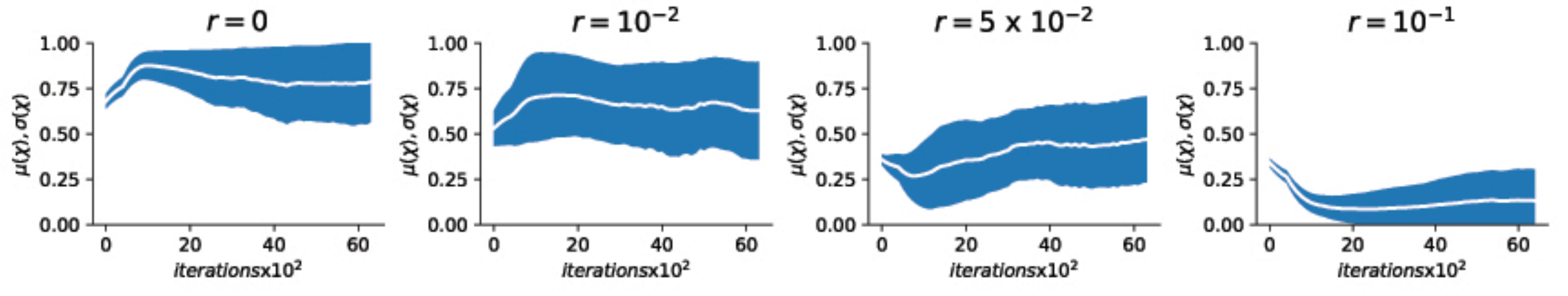}
    \caption{\small Different levels of regularization on the mean $\mu(\chi)$ (white line) and standard deviation $\sigma(\chi)$ (envelope) of the correction factors during training. Small regularization values force the correction factors to diverge faster. Increasing regularization collapses the correction factors hurting the message passing.}
\label{fig:gate_regularization}
\end{figure*}

\textbf{Patch Learning.} Tab. \ref{tab:ablation:bp4d} and \ref{tab:ablation:disfa} show results of AU-wise fusion for BP4D and DISFA (PP+F). On both, patch learning through fusion is beneficial, but on DISFA benefits are higher. This might be due to the fact that prediction results on DISFA are considerably more balanced across patches. Overall on BP4D the fusion improves results on almost all AUs compared to face prediction. This shows that even though the other patches perform worse on certain classes, there is structure to learn from their prediction that helps to improve performance. However, the fusion is not capable to replicate the result of the mouth prediction on AU14. On DISFA, in almost every case fusion gets close or higher to the best patch prediction. In both cases, fusion has greater problems in improving individual patches in cases where input predictions are already very noisy.

\begin{table*}[ht!]
    \centering\scriptsize
    \begin{tabular}{|c|c|c|c|c|c|c|c|c|c|}
        \hline   
        method & AU01 & AU02 & AU04 & AU06 & AU09 & AU12 & AU25 & AU26 & avg \\ \hline
        PP(right eye)  & 27.2 & 15.4  & 58.8 &  8.0 & 18.2 & 53.6 & 73.3 &  9.1  & 33.0  \\\hline
        PP(between eye)  & 34.6 & 13.2  & 59.7 & 15.4 & 21.1 & 50.9 & 72.9 &  8.5  & 34.5  \\\hline
        PP(mouth) &  7.5 & 6.4   & 44.6 & 28.5 & 23.9 & \textbf{72.1} & 87.5 & [27.3]  & 37.2  \\\hline  
        PP(right cheek) & 24.6 & 12.2 & 46.1 & 31.2 & 45.2 & 71.5 & 84.5 & 22.4  & 33.8 \\\hline
        PP(nose)  & 21.9 & 19.1  & 52.0 & \textbf{32.0} & \textbf{50.9} & 66.5 & 76.6 &  8.9  & 41.0  \\\hline
        PP(face) & 29.8 & [31.4]  & 64.6 & 26.8 & 21.3 & 70.1 & 87.0 & 20.3  & 43.9  \\\hline   
        PP+F      & [40.1] & 18.6 & \textbf{70.8} & 25.4 & 42.1 &  [71.8] & [88.8] &  26.4 & [48.0]  \\\hline
        DSIN & \textbf{42.4} & \textbf{39.0} & [68.4] & [28.6] & [46.8] & 70.8 & \textbf{90.4} & \textbf{42.2} & \textbf{53.6} \\\hline   
     \end{tabular}
     \caption{\small Results of DSIN on DISFA. PP([patch]) stands for patch prediction on the indicated patch. F stands for the fusion. DSIN is the final model. For DISFA we only show the DSIN with $T=10$, the best performing on BP4D.}
    \label{tab:ablation:disfa} 
\end{table*} 

\textbf{Structure Learning.} Tab. \ref{tab:ablation:bp4d} and \ref{tab:ablation:disfa} show results of the final DSIN model. For BP4D, we also perform a study of the number of iterations $T$ considered for structure inference. Since parameters $\omega_j$ are shared across iterations, more iterations are beneficial to capture AU relationships in a fully connected graph with a large number of nodes (12 in our case). We also trained DSIN without correction factors (Eq. \ref{eq:corrected_message} is not applied in this case). Results are inferior compared with the same model with correction factors. In the case of DISFA, we only applied the structure inference with the best previously found $T=10$ steps. Structure inference is beneficial in both cases. On BP4D, it considerably improves AU2 and AU14. For DISFA, the results are even more conclusive. Adding the structure inference brings more than 5\% improvement over the fusion. 

\textbf{Correction factor regularization.} Fig. \ref{fig:gate_regularization} shows the effect of increasing regularization applied on the correction factors $\chi$. Overall, regularizing $\chi$ does not bring significant benefits. When comparing $r=10^{-2}$ with no regularization the differences are minimal. The network has the ability to learn sparse message passing by itself without regularization. Still, small values of $r$ lead to faster divergence of $\chi$ and faster convergence of the network. The difference in performance is not significant. On the other hand values of $r>5 \times 10^{-2}$ negatively affect performance as most of $\chi$ get closer to $0$ and no messages are passed anymore. For these reasons, we keep $r=5 \times 10^{-3}$.

\begin{figure}[ht!]
\centering
\includegraphics[width=\linewidth]{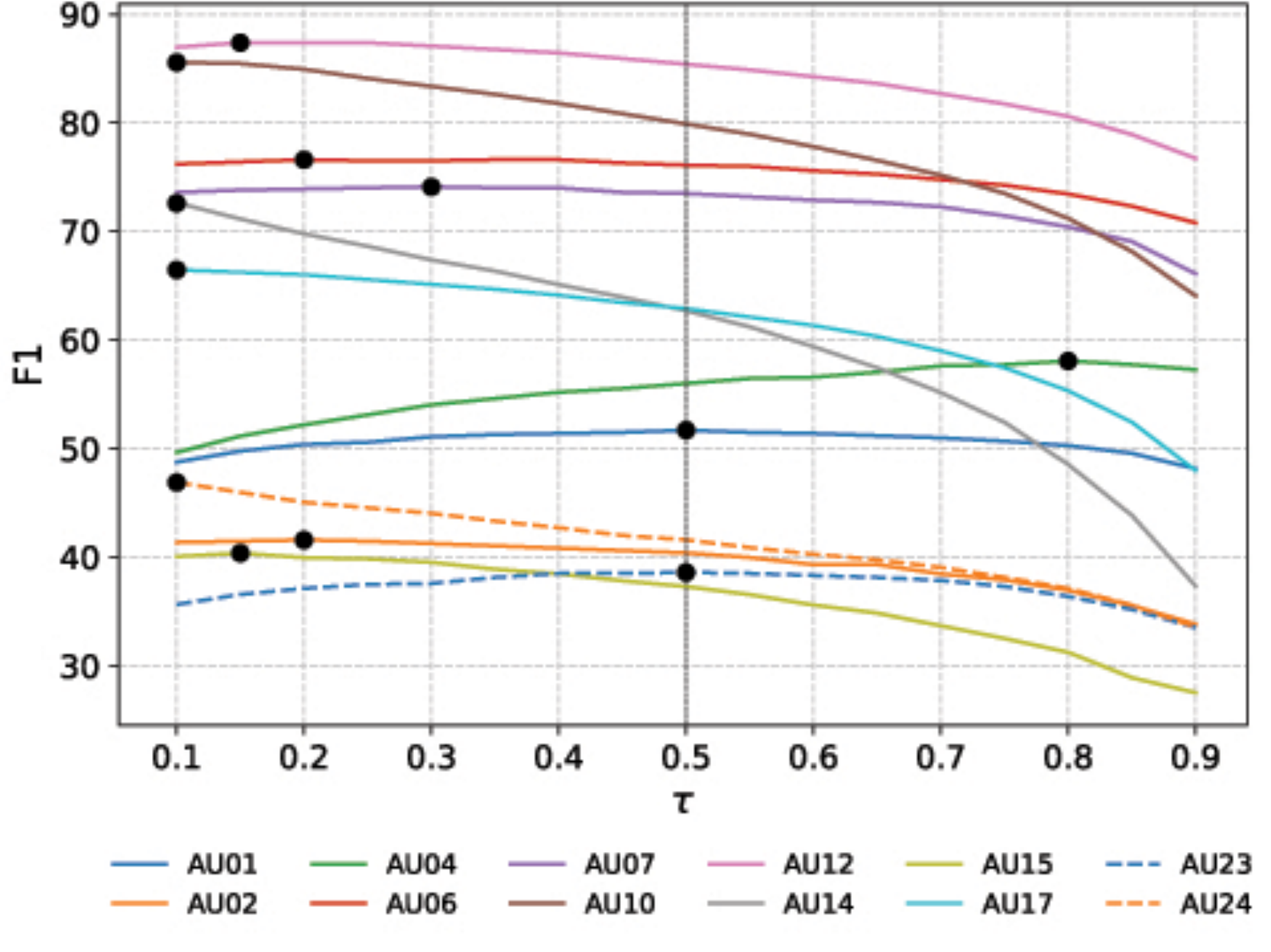}
\captionof{figure}{\small $\tau$ vs AU performance on BP4D validation set. Black circles denote best score.}
\label{fig:tune_t}
\end{figure}

\textbf{Threshold Tuning.} Prediction value per AU takes values between 0 and 1. In all results, we compute the performance by binarizing the output with respect to threshold $\tau=0.5$. Although class balancing as a weighted loss is beneficiary, it does not totally solve data imbalance. Fig. \ref{fig:tune_t} shows performance in terms of $\tau$ for validation set of BP4D. As shown, a threshold $\tau=0.5$ is not an ideal value. For most classes $\tau \in [0.1,0.3]$ is preferable. Exception is AU04. Tables \ref{tab:ablation:bp4d} and \ref{tab:ablation:disfa} show the performance of the proposed model after tuning $\tau$ per class (DSIN$^{tt}$). This way $2.8\%$ and $3.1\%$ of performance is gained on BP4D and DISFA, respectively. 

\begin{table*}[ht!]
    \centering
    \resizebox{\textwidth}{!}{%
    \begin{tabular}{|c|c|c|c|c|c|c|c|c|c|c|c|c|c|}
    \hline
    method & AU01 & AU02 & AU04 & AU06 & AU07 & AU10 & AU12 & AU14 & AU15 & AU17 & AU23 & AU24 & AVG\\ \hline
    JPML \cite{zhao2015joint} & 32.6 & 25.6 & 37.4 & 42.3 & 50.5 & 72.2 & 74.1 & [65.7] & 38.1 & 40.0 & 30.4 & [42.3] & 45.9 \\ \hline
    DRML \cite{zhao2016deep} & 36.4 & \textbf{41.8} & 43.0 & 55.0 & 67.0 & 66.3 & 65.8 & 54.1 & 33.2 & 48.0 & 31.7 & 30.0 & 48.3 \\ \hline
    CPM \cite{zeng2015confidence} & [43.4] & 40.7 & 43.3 & 59.2 & 61.3 & 62.1 & 68.5 & 52.5 & 36.7 & 54.3 & \textbf{39.5} & 37.8 & 50.0 \\ \hline
    ROI \cite{li2017action} & 36.2 & 31.6 & 43.4 & \textbf{77.1} & [73.7] & [85.0] & [87.0] & 62.6 & \textbf{45.7} & 58.0 & 38.3 & 37.4 & 56.4 \\ \hline
    DSIN & \textbf{51.7} & 40.4 & [56.0] & 76.1 & 73.5 & 79.9 & 85.4 & 62.7 & 37.3 & [62.9] & [38.8] & 41.6 & [58.9] \\\hline
    DSIN$^{tt}$ & \textbf{51.7} & [41.6] & \textbf{58.1} & [76.6] & \textbf{74.1} & \textbf{85.5} & \textbf{87.4} & \textbf{72.6} & [40.4] & \textbf{66.5} & 38.6 & \textbf{46.9} & \textbf{61.7}
    \\ \hline 
    \end{tabular}}
    \caption{\small AU recognition results on BP4D. Best results are shown in bold. Second best results are shown in brackets. For the proposed model we show an additional set of results (DSIN$_{tt}$) obtained when the decision threshold is tuned per AU.}
    \label{tab:sota:bp4d}
\end{table*}

\begin{table*}[ht!]
    \centering\scriptsize
    \begin{tabular}{|c|c|c|c|c|c|c|c|c|c|c|c|c|c|}
    \hline
        method & AU01 & AU02 & AU04 & AU06 & AU09 & AU12 & AU25 & AU26 & avg \\ \hline
        APL\cite{zhong2015learning} & 11.4 & 12.0 & 30.1 & 12.4 & 10.1 & 65.9 & 21.4 & 26.0 & 23.8 \\ \hline
        DRML \cite{zhao2016deep} & 17.3 & 17.7 & 37.4 & 29.0 & 10.7 & 37.7 & 38.5 & 20.1 & 26.7 \\ \hline
        ROI \cite{li2017action} & 41.5 & 26.4 & 66.4 & \textbf{50.7} & 8.5 & \textbf{89.3} & 88.9 & 15.6 & 48.5 \\ \hline
        DSIN & [42.4] & [39.0] & [68.4] & 28.6 & [46.8] & 70.8 & [90.4] & [42.2] & [53.6] \\\hline  
        DSIN$^{tt}$ & \textbf{46.9} & \textbf{42.5} & \textbf{68.8} & [32.0] & \textbf{51.8} & [73.1] & \textbf{91.9} & \textbf{46.6} & \textbf{56.7} \\\hline 
    \end{tabular}
    \caption{\small AU recognition results on DISFA. Best results are shown in bold. Second best results are shown in brackets.}
    \label{tab:sota:disfa}
\end{table*}

\subsubsection{Comparison with state-of-the-art.}
\label{sec:experimental_analysis:sota}

Tables \ref{tab:sota:bp4d} and \ref{tab:sota:disfa} show how our model compares against the state-of-the-art related methods on BP4D and DISFA, respectively. DSIN and ROI are the best performing in both datasets. Both methods learn deep local representations and patch combinations end-to-end. The worst performing methods, JPML on BP4D and APL on DISFA, use predefined features and are not end-to-end trained. Comparing DSIN and ROI with DRML one can observe the advantage in learning independent local representation. Both ROI and our model learn independent local representations, while DRML disentangles the representation learning in just one layer of their network. Interestingly though, there is also an exception. On BP4D, CPM performs slightly better than DRML even though it is not a deep learning method. 
When comparing our proposed model with ROI on BP4D our CNN trained just on face without class balancing has inferior results. When we include class balancing and patch learning our topology improves performance, further enhanced by structure inference and end-to-end final training. In the case of DISFA, single CNN trained on the whole face with class balancing has a performance of 43.9, being $4.6\%$ lower than ROI. When we add patch prediction fusion (PP+F) we get just $0.5\%$ lower than ROI while the addition of the structure inference and threshold tuning improves ROI performance. Finally, DSIN shows the best results on both datasets. For BP4D, from the 12 AUs target it performs best on 5 and second best on additional 5. In the case of DISFA the improvement over ROI is greater, DSIN performing best in all but one AU. Overall, we obtain 5.3\% absolute and 9.4\% relative performance improvement on BP4D and 8.2\% absolute and 16.9\% relative performance improvement on DISFA, respectively. 

\begin{figure*}[t]
    \centering
    \includegraphics[width=\textwidth]{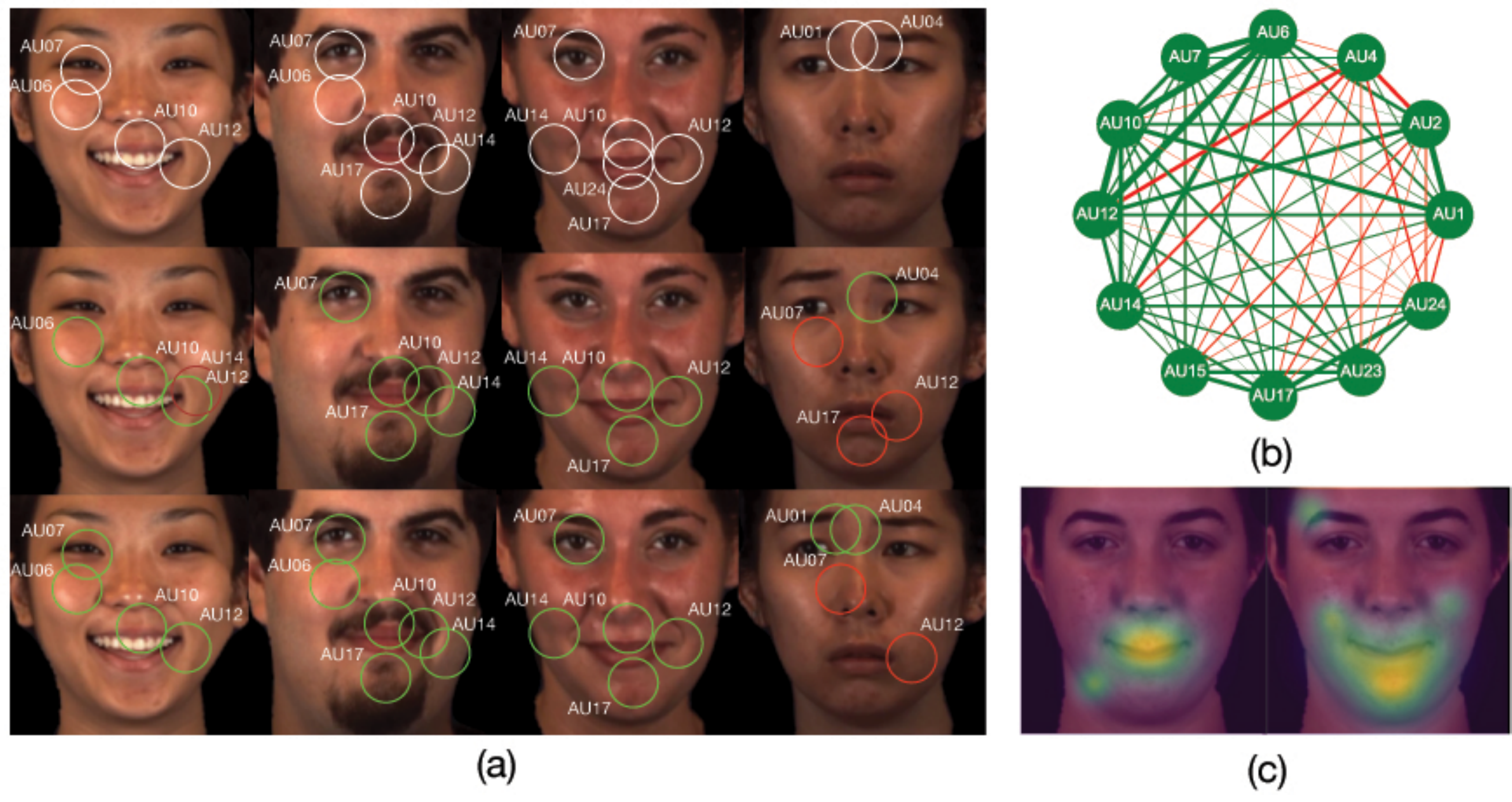}
    \caption{\small (a) Examples of AU predictions: ground-truth (top), fusion module (middle) and structure inference (bottom) prediction (\textcolor{green}{$\bullet$}: true positive, \textcolor{red}{$\bullet$}: false positive). (b) AUs correlation in BP4D (\textcolor{green}{$\bullet$}: positive, \textcolor{red}{$\bullet$}: negative). Line thickness is proportional with correlation magnitude. (c) Class activation map for AU24 that shows the discriminative regions of simple patch prediction (left) and DSIN (right). Best seen in color.}
    \label{fig:qualitative}
\end{figure*}

\subsubsection{Qualitative results.}
\label{sec:experimental_analysis:qualitative}
Fig. \ref{fig:qualitative}(a) shows some examples of how structure inference tends to correct predictions following AU correlations. We show the magnitude of AU correlations on BP4D in Fig. \ref{fig:qualitative}(b). In the first 3 column examples, AU06 and AU07 are not correctly classified by the fusion model (middle row). Both these AUs are highly correlated with already detected AUs like AU10, AU12 and AU14. Such correlation could be captured by SI (bottom row). The rightmost example shows how AU17, a false positive, is corrected. As shown in Fig. \ref{fig:qualitative}(b), AU17 is negatively correlated with AU4, which was already detected. In Fig. \ref{fig:qualitative}(c) we show a class activation map \cite{selvaraju2016grad} for AU24 of the patch prediction (left) vs. the DSIN (right). Contrary to very localized patch prediction, the attention on right expands to a larger area of the face where possible correlated AUs might exist, e.g. AU15, AU17 and AU23.

\section{Conclusion}
\label{sec:conclusion}

We proposed the Deep Structured Inference Network (DSIN), a deep network designed to deal with patch and multi-label learning for AU recognition in an integrated way. DSIN first learns independent deep local and global representations and corresponding predictions. Then, it learns relationships between predictions per AU through stacked fully connected layers. Finally, inspired by inference algorithms in graphical models, DSIN replicates a message passing mechanism in a connectionist fashion. This adds the ability to capture correlations in the output space. The model is end-to-end trainable and improves state-of-the-art results by 5.3\% and 8.2\% performance on BP4D and DISFA datasets, respectively. Future work includes learning patch structure at feature level and a structure inference module with increased capacity for output structure learning.       


{\small
\bibliographystyle{ieee}
\bibliography{egbib}

\begin{thebibliography}{10}\itemsep=-1pt

\bibitem{bakkes2012personalised}
S.~Bakkes, C.~T. Tan, and Y.~Pisan.
\newblock Personalised gaming.
\newblock {\em JCT}, 3, 2012.

\bibitem{chu2016crf}
X.~Chu, W.~Ouyang, X.~Wang, et~al.
\newblock Crf-cnn: Modeling structured information in human pose estimation.
\newblock In {\em Advances in Neural Information Processing Systems}, pages
  316--324, 2016.

\bibitem{deng2016structure}
Z.~Deng, A.~Vahdat, H.~Hu, and G.~Mori.
\newblock Structure inference machines: Recurrent neural networks for analyzing
  relations in group activity recognition.
\newblock In {\em Proceedings of the IEEE Conference on Computer Vision and
  Pattern Recognition}, pages 4772--4781, 2016.

\bibitem{devault2014}
D.~DeVault, R.~Artstein, G.~Benn, T.~Dey, E.~Fast, A.~Gainer, and L.-P.
  Morency.
\newblock A virtual human interviewer for healthcare decision support.
\newblock {\em AAMAS}, 2014.

\bibitem{ding2013facial}
X.~Ding, W.-S. Chu, F.~De~la Torre, J.~F. Cohn, and Q.~Wang.
\newblock Facial action unit event detection by cascade of tasks.
\newblock In {\em Proceedings of the IEEE International Conference on Computer
  Vision}, pages 2400--2407, 2013.

\bibitem{ekman2002facial}
P.~Ekman, W.~Friesen, and J.~Hager.
\newblock Facs manual. a human face.
\newblock 2002.

\bibitem{eleftheriadis2015multi}
S.~Eleftheriadis, O.~Rudovic, and M.~Pantic.
\newblock Multi-conditional latent variable model for joint facial action unit
  detection.
\newblock In {\em Proceedings of the IEEE International Conference on Computer
  Vision}, pages 3792--3800, 2015.

\bibitem{benitez2017recognition}
C.~Fabian Benitez-Quiroz, Y.~Wang, and A.~M. Martinez.
\newblock Recognition of action units in the wild with deep nets and a new
  global-local loss.
\newblock In {\em The IEEE International Conference on Computer Vision (ICCV)},
  Oct 2017.

\bibitem{frith2009role}
C.~Frith.
\newblock Role of facial expressions in social interactions.
\newblock {\em Philosophical Transactions of the Royal Society B: Biological
  Sciences}, 364(1535):3453--3458, 2009.

\bibitem{jaiswal2016deep}
S.~Jaiswal and M.~Valstar.
\newblock Deep learning the dynamic appearance and shape of facial action
  units.
\newblock In {\em Applications of Computer Vision (WACV), 2016 IEEE Winter
  Conference on}, pages 1--8. IEEE, 2016.

\bibitem{kaltwang2012continous}
S.~Kaltwang, O.~Rudovic, and M.~Pantic.
\newblock Continuous pain intensity estimation from facial expressions.
\newblock {\em ISVC}, pages 368--377, 2012.

\bibitem{kapoor2007automatic}
A.~Kapoor, W.~Burleson, and R.~W. Picard.
\newblock Automatic prediction of frustration.
\newblock {\em IJHCS}, 65(8):724--736, 2007.

\bibitem{kazemi2014one}
V.~Kazemi and J.~Sullivan.
\newblock One millisecond face alignment with an ensemble of regression trees.
\newblock In {\em Proceedings of the IEEE Conference on Computer Vision and
  Pattern Recognition}, pages 1867--1874, 2014.

\bibitem{kulkarni2017automatic}
K.~Kulkarni, C.~A. Corneanu, I.~Ofodile, S.~Escalera, X.~Baro, S.~Hyniewska,
  J.~Allik, and G.~Anbarjafari.
\newblock Automatic recognition of deceptive facial expressions of emotion.
\newblock {\em arXiv preprint arXiv:1707.04061}, 2017.

\bibitem{li2017action}
W.~Li, F.~Abitahi, and Z.~Zhu.
\newblock Action unit detection with region adaptation, multi-labeling learning
  and optimal temporal fusing.
\newblock {\em arXiv preprint arXiv:1704.03067}, 2017.

\bibitem{lucey10}
P.~Lucey, J.~F. Cohn, I.~Matthews, S.~Lucey, S.~Sridharan, J.~Howlett, and
  K.~M. Prkachin.
\newblock Automatically detecting pain in video through facial action units.
\newblock {\em SMC-B}, 41(3):664--674, 2011.

\bibitem{mavadati2013disfa}
S.~M. Mavadati, M.~H. Mahoor, K.~Bartlett, P.~Trinh, and J.~F. Cohn.
\newblock Disfa: A spontaneous facial action intensity database.
\newblock {\em IEEE Transactions on Affective Computing}, 4(2):151--160, 2013.

\bibitem{parkhi2015deep}
O.~M. Parkhi, A.~Vedaldi, and A.~Zisserman.
\newblock Deep face recognition.
\newblock In {\em British Machine Vision Conference}, 2015.

\bibitem{selvaraju2016grad}
R.~R. Selvaraju, M.~Cogswell, A.~Das, R.~Vedantam, D.~Parikh, and D.~Batra.
\newblock Grad-cam: Visual explanations from deep networks via gradient-based
  localization.
\newblock {\em See https://arxiv. org/abs/1610.02391 v3}, 7(8), 2016.

\bibitem{simonyan2014very}
K.~Simonyan and A.~Zisserman.
\newblock Very deep convolutional networks for large-scale image recognition.
\newblock {\em arXiv preprint arXiv:1409.1556}, 2014.

\bibitem{song2015exploiting}
Y.~Song, D.~McDuff, D.~Vasisht, and A.~Kapoor.
\newblock Exploiting sparsity and co-occurrence structure for action unit
  recognition.
\newblock In {\em Automatic Face and Gesture Recognition (FG), 2015 11th IEEE
  International Conference and Workshops on}, volume~1, pages 1--8. IEEE, 2015.

\bibitem{taigman2014deepface}
Y.~Taigman, M.~Yang, M.~Ranzato, and L.~Wolf.
\newblock Deepface: Closing the gap to human-level performance in face
  verification.
\newblock In {\em Proceedings of the IEEE conference on computer vision and
  pattern recognition}, pages 1701--1708, 2014.

\bibitem{vinciarelli09}
A.~Vinciarelli, M.~Pantic, and H.~Bourlard.
\newblock Social signal processing: Survey of an emerging domain.
\newblock {\em IVC}, 27(12):1743--1759, 2009.

\bibitem{walecki2017deep}
R.~Walecki, V.~Pavlovic, B.~Schuller, M.~Pantic, et~al.
\newblock Deep structured learning for facial action unit intensity estimation.
\newblock {\em arXiv preprint arXiv:1704.04481}, 2017.

\bibitem{wang2013capturing}
Z.~Wang, Y.~Li, S.~Wang, and Q.~Ji.
\newblock Capturing global semantic relationships for facial action unit
  recognition.
\newblock In {\em Proceedings of the IEEE International Conference on Computer
  Vision}, pages 3304--3311, 2013.

\bibitem{wu2016constrained}
Y.~Wu and Q.~Ji.
\newblock Constrained joint cascade regression framework for simultaneous
  facial action unit recognition and facial landmark detection.
\newblock In {\em Proceedings of the IEEE Conference on Computer Vision and
  Pattern Recognition}, pages 3400--3408, 2016.

\bibitem{zeng2015confidence}
J.~Zeng, W.-S. Chu, F.~De~la Torre, J.~F. Cohn, and Z.~Xiong.
\newblock Confidence preserving machine for facial action unit detection.
\newblock In {\em Proceedings of the IEEE International Conference on Computer
  Vision}, pages 3622--3630, 2015.

\bibitem{zhang2016task}
X.~Zhang and M.~H. Mahoor.
\newblock Task-dependent multi-task multiple kernel learning for facial action
  unit detection.
\newblock {\em Pattern Recognition}, 51:187--196, 2016.

\bibitem{zhang2014bp4d}
X.~Zhang, L.~Yin, J.~F. Cohn, S.~Canavan, M.~Reale, A.~Horowitz, P.~Liu, and
  J.~M. Girard.
\newblock Bp4d-spontaneous: a high-resolution spontaneous 3d dynamic facial
  expression database.
\newblock {\em Image and Vision Computing}, 32(10):692--706, 2014.

\bibitem{zhao2015joint}
K.~Zhao, W.-S. Chu, F.~De~la Torre, J.~F. Cohn, and H.~Zhang.
\newblock Joint patch and multi-label learning for facial action unit
  detection.
\newblock In {\em Proceedings of the IEEE Conference on Computer Vision and
  Pattern Recognition}, pages 2207--2216, 2015.

\bibitem{zhao2016deep}
K.~Zhao, W.-S. Chu, and H.~Zhang.
\newblock Deep region and multi-label learning for facial action unit
  detection.
\newblock In {\em Proceedings of the IEEE Conference on Computer Vision and
  Pattern Recognition}, pages 3391--3399, 2016.

\bibitem{zheng2015conditional}
S.~Zheng, S.~Jayasumana, B.~Romera-Paredes, V.~Vineet, Z.~Su, D.~Du, C.~Huang,
  and P.~H. Torr.
\newblock Conditional random fields as recurrent neural networks.
\newblock In {\em Proceedings of the IEEE International Conference on Computer
  Vision}, pages 1529--1537, 2015.

\bibitem{zhong2015learning}
L.~Zhong, Q.~Liu, P.~Yang, J.~Huang, and D.~N. Metaxas.
\newblock Learning multiscale active facial patches for expression analysis.
\newblock {\em IEEE transactions on cybernetics}, 45(8):1499--1510, 2015.

\end{thebibliography}
}

\end{document}